\documentclass[preprint,5p,times,twocolumn]{elsarticle}

\usepackage{amssymb}

\usepackage{xspace}
\usepackage{amsmath,bm}
\usepackage{multirow}
\usepackage{multicol}
\usepackage{siunitx}
\usepackage{pifont}
\usepackage{soul}
\usepackage{colortbl}
\usepackage[table,x11names,dvipsnames]{xcolor}
\usepackage{pgfplots}
\usepackage{caption}
\usepackage{tabularx}
\usepackage{subcaption}
\usepackage{booktabs}
\usepackage[colorlinks = true,
            linkcolor = blue,
            urlcolor  = blue,
            citecolor = green,
            anchorcolor = blue]{hyperref}

\newcommand{\cmark}{\ding{51}}
\newcommand{\xmark}{\ding{55}}

\newcommand*{\eg}{e.g.\@\xspace}
\newcommand*{\ek}{EPIC-KITCHENS-100\@\xspace}
\newcommand*{\ie}{i.e.\@\xspace}
\newcommand*{\ours}{SeqDG\@\xspace}
\newcommand*{\oursmix}{SeqMix\@\xspace}

\definecolor{up_color}{RGB}{204,51,0}
\definecolor{down_color}{RGB}{202,195,121}

\definecolor{cadmiumgreen}{rgb}{0.0, 0.42, 0.24}
\newcommand{\up}{{\color{cadmiumgreen}$\blacktriangle$}\hspace{.3em}}

\newcommand{\revised}{\textcolor{black}}
\newcommand{\secondrevised}{\textcolor{black}}

\journal{Pattern Recognition Letters}

\usepackage{lipsum}

\newcommand\blfootnote[1]{%
  \begingroup
  \renewcommand\thefootnote{}\footnote{#1}%
  \addtocounter{footnote}{-1}%
  \endgroup
}

\begin{document}

\begin{frontmatter}









\title{Domain Generalization using Action Sequences\\for Egocentric Action Recognition} 

\author[0]{Amirshayan Nasirimajd} 
\ead{amirshayan.nasirimajd@studenti.polito.it}

\author[0]{Chiara Plizzari}  
\ead{chiara.plizzari@polito.it}

\author[0]{Simone Alberto Peirone}
\ead{simone.peirone@polito.it}

\author[0]{Marco Ciccone}
\ead{marco.Ciccone@polito.it}

\author[0]{Giuseppe Averta}
\ead{giuseppe.averta@polito.it}

\author[0]{Barbara Caputo}
\ead{barbara.caputo@polito.it}

            
\affiliation[0]{organization={Politecnico di Torino, Department of Control and Computer Engineering},
            addressline={Corso Castelfidardo, 34/d}, 
            city={Turin},
            postcode={10138}, 
            state={TO},
            country={Italy}}

\begin{abstract}
Recognizing human activities from visual inputs, particularly through a first-person viewpoint, is essential for enabling robots to replicate human behavior. Egocentric vision, characterized by cameras worn by observers, captures diverse changes in illumination, viewpoint, and environment. This variability leads to a notable drop in the performance of Egocentric Action Recognition models when tested in environments not seen during training.
In this paper, we tackle these challenges by proposing a domain generalization approach for Egocentric Action Recognition. Our insight is that action sequences often reflect consistent user intent across visual domains. By leveraging \textit{action sequences}, we aim to enhance the model's generalization ability across unseen environments. Our proposed method, named \ours, introduces a visual-text sequence reconstruction objective (SeqRec) that uses contextual cues from both text and visual inputs to reconstruct the central action of the sequence. Additionally, we enhance the model's robustness by training it on mixed sequences of actions from different domains (SeqMix).
We validate \ours on the EGTEA and
EPIC-KITCHENS-100 datasets. Results \revised{on EPIC-KITCHENS-100,} show that \ours leads to +2.4\% \secondrevised{relative} average improvement in \revised{cross-domain} action recognition in unseen environments, 
\revised{and on EGTEA the model achieved +0.6\% Top-1 accuracy over SOTA in intra-domain action recognition.}
 Code and data: \href{https://github.com/Ashayan97/SeqDG}{Github.com/Ashayan97/SeqDG}
\end{abstract}







\end{frontmatter}
\blfootnote{This study was carried out within the FAIR - Future Artificial Intelligence Research and received funding from the European Union Next-GenerationEU (PIANO NAZIONALE DI RIPRESA E RESILIENZA (PNRR) – MISSIONE 4 COMPONENTE 2, INVESTIMENTO 1.3 – D.D. 1555 11/10/2022, PE00000013). This manuscript reflects only the authors’ views and opinions, neither the European Union nor the European Commission can be considered responsible for them. 

This study also acknowledge the CINECA award under the ISCRA initiative, for the availability of high performance computing resources and support.}

\section{Introduction}

Egocentric vision offers a unique (first-person) perspective to capture human activities, which serves as a foundational tool in many fields, including assistive technologies~\cite{ohnbar2018personalized} and robotics applications~\cite{park2016egocentric}.
Among many egocentric vision tasks~\cite{plizzari2024outlook}, Egocentric Action Recognition (EAR) represents a crucial component to build comprehensive models of human actions.

A major challenge in deploying action recognition models in the real-world lies in their ability to 
generalize to environments and scenarios that significantly differ from those seen during training (\emph{domain shift})~\cite{plizzari2023can,peirone2025egocentric}. Indeed, EAR models often become overly dependent on specific environmental cues present in the training data\secondrevised{~\cite{kowal2022deeper}, which could limit their ability to recognize actions when performed in new, unseen environments~\cite{torralba2011unbiased}}. 
\begin{figure}[t]
    \centering
    \includegraphics[width=.9\columnwidth]{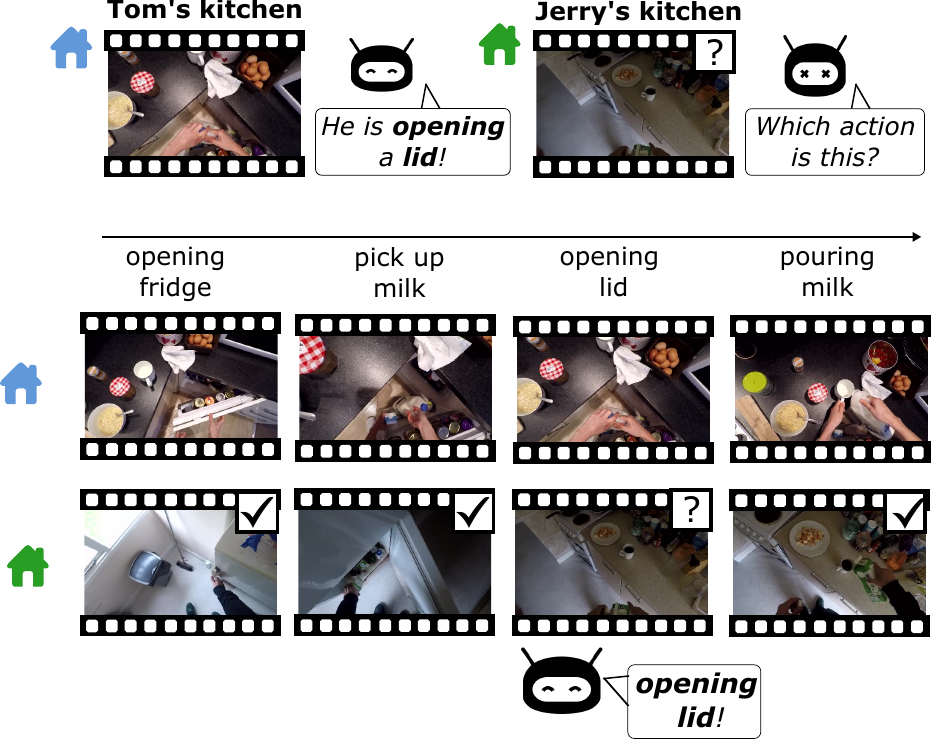}
    \caption{Egocentric Action Recognition (EAR) models struggle in classifying the same action when it is performed in different environments (top). Despite different visual contexts, the sequence of actions — opening the fridge, picking up milk, opening lid and pouring milk — remains consistent. Thus, considering previous and subsequent actions within a sequence proves beneficial for improving cross-domain robustness of the EAR models.
    }
    
    \label{fig:teaser}
\end{figure}

\begin{figure}[t]
    \centering
    \includegraphics[width=1\columnwidth, trim={0 0 0 0}]{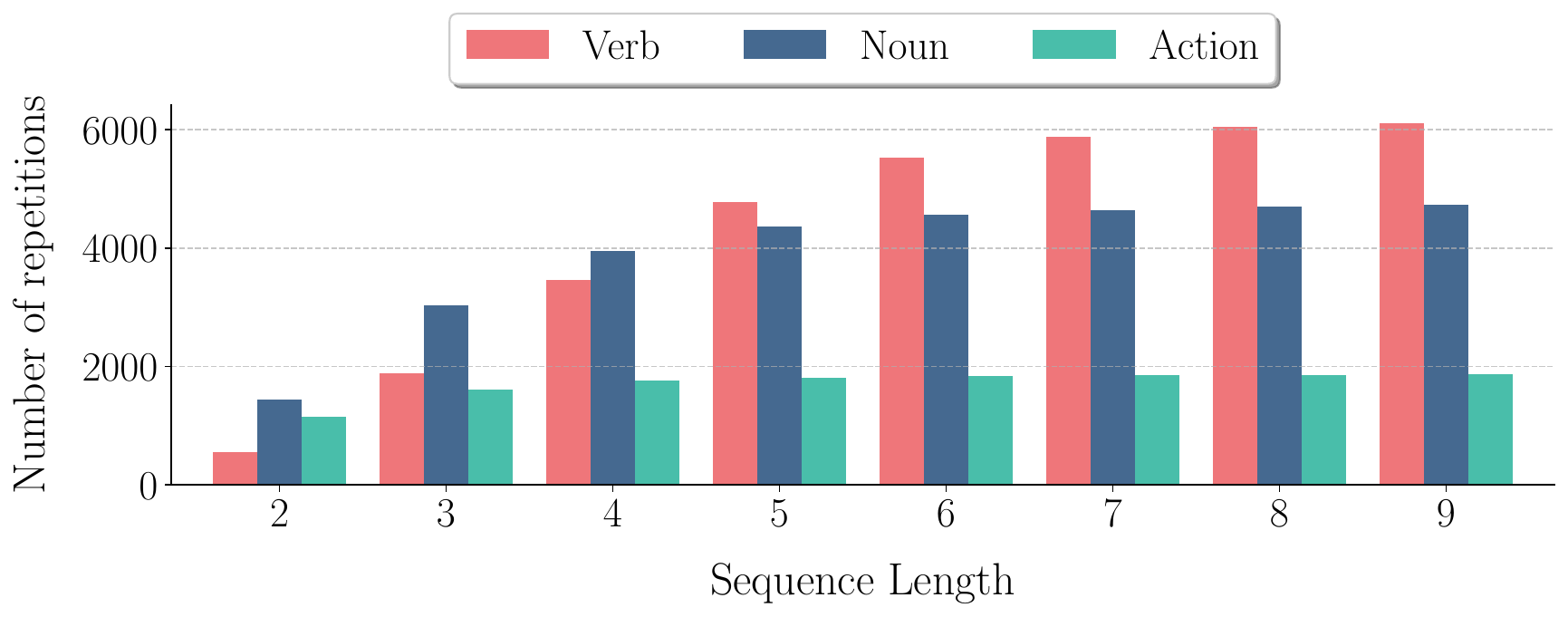}
    \caption{Number of repeated Verb, Noun and Action sequences in the \ek dataset. For each sequence length, we report the number of sub-sequences of any smaller (or equal) length that repeat across domains.}
    \label{fig:hist_seqnumber}
\end{figure}

Moreover, unique behavioral traits may influence the way similar actions are performed and add an extra layer of complexity in action recognition. A key insight we can leverage to improve generalization is that human actions do not occur in isolation but sequentially and are naturally goal-driven~\cite{cooper2006hierarchical}, making action sequences crucial to understanding individual actions~\cite{kazakos_temporal2021}.
By exploiting the additional context provided by surrounding actions, we can in turn reduce the impact of confounding factors, such as changes in the environment or user behavior.
\secondrevised{As an example, consider} two people preparing an omelet, each in their own kitchen, generally perform the same steps, one after the other: they take eggs from the refrigerator, break them, add spices or other ingredients, and finally fry the omelet. 
\secondrevised{Kitchens and users may be different, but the steps needed to prepare the omelet are the same.}
A similar example is shown in  Fig.~\ref{fig:teaser}: despite different contexts, the sequence of actions -- opening the fridge, picking up milk, opening the lid, and pouring milk -- are the same. 
We validate these observations by looking at the number of repeated action sequences 
in one of the most significant egocentric vision datasets, EPIC-KITCHENS-100~\cite{Damen2022RESCALING}.
Samples are annotated with a \textit{Verb} and a \textit{Noun} label, as well as their combination, indicated as \textit{Action}.
Actions are recorded in 45 different kitchens, by 37 different participants.
Despite the large diversity in environments and participants, we observe a significant number of repeated action sequences: out of 32k samples, 5882, 4649, and 1852 sequences of length up to 5 are repeated for the Verb, Noun, and Action categories, respectively (Fig.~\ref{fig:hist_seqnumber}).

In this work, we claim that \textit{action sequences} can help generalization in two ways: (i)
although the specific execution can differ -- such as how eggs are cracked or ingredients are mixed -- the overall sequence of actions forms consistent patterns across different subjects; (ii) 
because these sequences are independent of the environment's layout or appearance, they can be used to adapt to and recognize individual actions in visually diverse scenarios, mitigating the negative effect of domain shift.
We hence propose \textbf{\ours}, a Domain Generalization method for EAR that leverages the similarity of action sequences across different visual domains to improve the generalization of action recognition models.
\ours integrates \textbf{SeqRec}, a visual-text sequence reconstruction objective that utilizes contextual information from both text and visual modalities to reconstruct a sequence's masked action. 
This serves as a \textit{proxy task} that explicitly pushes the model to capture temporal dependencies and relational cues across the sequence.
In addition, we introduce \textbf{SeqMix}, a data augmentation technique that mixes actions that share the same label but come from different domains to make the model more robust to visual changes. 

We note the distinction of Domain Generalization~(DG) from the Unsupervised Domain Adaptation~(UDA) setting, where unlabeled target samples are available during training~\cite{munro2020multi,song2021spatio,kim2021learning}.
We do not only demonstrate the advantages of our approach for cross-domain generalization but also outperform standard \secondrevised{UDA} methods and advanced sequence-based solutions, despite not accessing target data during training.

In summary, we present the following contributions:
\begin{itemize}
  \setlength\itemsep{0em}
    \item we highlight the role of action sequences in cross-domain EAR, guided by the principle that human actions follow patterns that are repeated across different environments;
    \item we propose \ours, a DG approach for EAR combining a visual-text reconstruction objective with mixed sequences across different domains, to encourage the model to better exploit action sequences;
    \item we validate \ours on EGTEA and EPIC-KITCHENS-100, observing SOTA performance and demonstrating that effective usage of sequence helps EAR in unseen domains.
\end{itemize}

\section{Related works}
\subsection{Egocentric Action Recognition}
CNN-based approaches for EAR can be divided into two main categories, using either 3D~\cite{carreira2017quo,feichtenhofer2019slowfast} or 2D~\cite{wang2016temporal,lin2019tsm} convolutions. 
\secondrevised{The goal is to adapt CNNs to the temporal nature of the videos, allowing for better modeling of the temporal context,
which is crucial for many tasks, \eg, action segmentation~\cite{wang2016temporal, quattrocchi2024synchronization} and action recognition~\cite{tu2022joint, liu2022motion}, specifically in EAR setting~\cite{kazakos_temporal2021}}.
Additionally, 
various approaches in EAR have explored integrating multiple modalities, a common strategy in video understanding to enhance model robustness~\cite{radevski2023multimodal, ramazanova2024exploring}. For instance, some works focus on the fusion of audio and visual information~\cite{kazakos2019epic, xiao2020audiovisual, chalk2024tim} \secondrevised{
or other non-audiovisual modalities, like IMU~\cite{zhang2024masked}. 
} 
\secondrevised{ They also improved the processing of different modalities—especially text—enabling the use of both vision and text to enhance long-term video understanding~\cite{sun2022long}.
}
In
\revised{most EAR} approaches, actions are classified independently, \ie, the temporal context from surrounding actions is not considered.
In contrast, sequence-based approaches~\cite{kazakos_temporal2021} integrate temporal context to improve action recognition.
\secondrevised{We design \ours on the intuition that temporal context may help EAR models, leveraging text to represent action sequences in a domain-agnostic way.}
\subsection{Domain Adaptation and Generalization}
%
Unsupervised Domain Adaptation (UDA) combines training on labeled source data with unlabeled target data to align their representations.
In adversarial approaches, alignment is achieved using adversarial classifiers.
Among these, TA3N~\cite{chen2019temporal} uses multiple adversarial branches to align the distributions at different temporal scales, while MM-SADA~\cite{munro2020multi} learns to align source and target distributions and features of different modalities. Similarly, CIA~\cite{yang2022interact} combines adversarial training with a gating mechanism to leverage cross-modal synergies.
Other works explored contrastive alignment across different modalities and domains~\cite{kim2021learning} or category-aware alignment~\cite{song2021spatio}.

Domain Generalization (DG) learns robust domain-agnostic representations without target data.
VideoDG~\cite{yao2021videodg} aligns local temporal features across domains as they generalize better compared to global features. RNA~\cite{planamente2024relative} introduces a cross-domain and cross-modality norm alignment loss to learn equally from all domains and modalities. \revised{EgoZAR~\cite{peirone2025egocentric} proposes leveraging domain-agnostic representations of the specific locations associated with particular actions (activity-centric zones) to enhance the generalization 
across different environments. CIR~\cite{plizzari2023can} implements cross-instance reconstruction of each video representation from other domains videos.}
%

Previous works in DG/UDA investigated EAR on individual actions.
Here, we study a new paradigm for DG in EAR that leverages action sequences, exploiting the observation that they repeat in different domains in accordance with the user' intentions, and may help EAR models in unseen environments.
\subsection{Masked Autoencoders for video applications}
Masked Autoencoders (MAEs)~\cite{he2022masked} learn to reconstruct missing parts of input data, which demonstrates to be an effective pre-training objective for various downstream tasks, with applications also to videos\secondrevised{~\cite{tong2022videomae, wang2023videomae}} and multi-modal vision-language inputs~\cite{geng2022multimodal}. 
Most related to our work is Voltron~\cite{karamcheti2023language}, which reconstructs video patches at the pixel level conditioned by textual narrations and generates captions for input clips.
\secondrevised{Unlike previous approaches, which reconstruct masked inputs at the  pixel level~\cite{tong2022videomae,wang2023videomae, geng2022multimodal,karamcheti2023language}, \ours is designed to reconstruct masked actions in a sequence to achieve high-level reasoning about human activities. }

\section{Proposed Method}
\begin{figure}
    \centering
    \includegraphics[width=1\columnwidth]{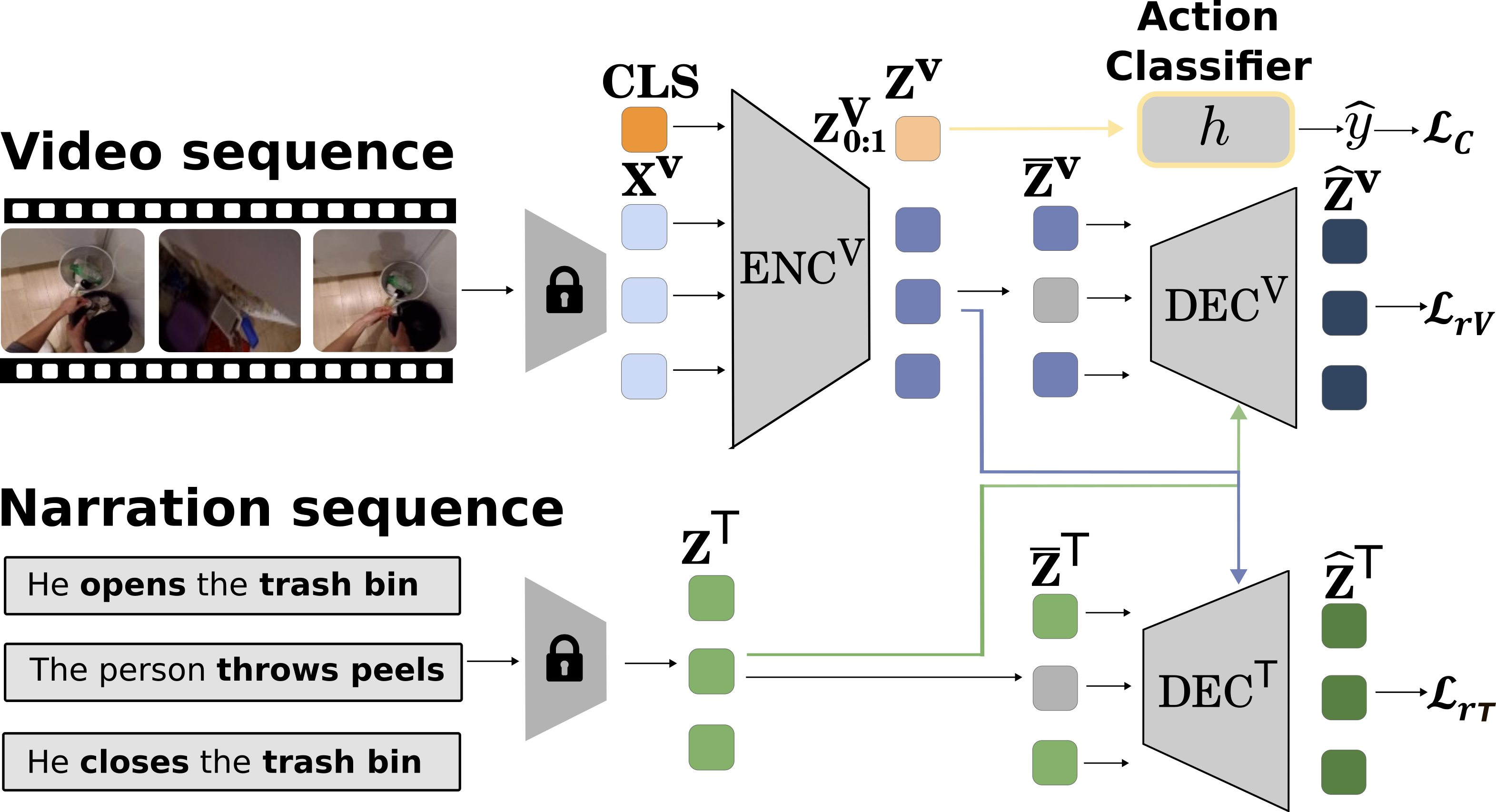}
    \caption{
    \textbf{\ours architecture.} 
    We are given visual and textual inputs $\bm{X}^V$ and $\bm{X}^T$. A classification token $\texttt{CLS}$ is appended to the visual input for classification. Visual inputs are fed to an encoder $\texttt{ENC}^\texttt{V}$, resulting in intermediate visual embeddings $\bm{Z}_i^V$, while textual features are passed through an identity function to get $\bm{Z}_i^T$. The latter are masked ($\bm{\bar{Z}}_i^V$ and $\bm{\bar{Z}}_i^T$) and fed to two separate decoders $\texttt{DEC}^\texttt{V}$ and $\texttt{DEC}^\texttt{T}$ for visual and text reconstruction ($\mathcal{L}_{rV}$ and $\mathcal{L}_{rT}$). The transformed $\texttt{CLS}$ token {$\bm{Z^v_{0:1}}$} is fed to classifier $h$ for action classification ($\mathcal{L}_{C}$). 
    }
    \label{fig:method}
\end{figure}

Our goal is to increase the robustness of action recognition models in the presence of domain shift, by leveraging the similarity of action sequences across different domains.
Relying solely on visual information might be problematic as visual data is susceptible to variations that may not be related to the actions themselves, e.g., different environments, backgrounds, lighting conditions, etc. 
In contrast, textual narrations are less tied to the domain and inherently more robust to domain shift~\cite{plizzari2023can}. 
We propose a visual-text sequence reconstruction objective (\textbf{SeqRec}, Section~\ref{sec:seqrec}), that reconstructs a masked action of the sequence using the information of surrounding actions from both the text and visual modalities.
Additionally, we train the network with sequences that combine actions from various domains (\textbf{SeqMix}, Section \ref{sec:seqmix}). This approach is designed to explicitly model different distributions of actions within sequences during training, thereby increasing performance on out-of-distribution test data. We call our overall approach \textbf{SeqDG}. 
Formally, we define \ours as a Domain Generalization approach for EAR, in which the model is trained on action samples from a set of $N_s$ different \textit{source} domains $\{\mathcal{D}_1^s,\dots,\mathcal{D}_{N_s}^s\}$ and evaluated on a set of $N_t$ unseen \textit{target} domains $\{\mathcal{D}_1^t,\dots,\mathcal{D}_{N_t}^t\}$, under the assumption that there is no overlap between the different domains.
An illustration of the overall method is shown in Figure \ref{fig:method}.

\subsection{Leveraging sequences of actions}
\secondrevised{\ours leverages sequences of actions rather than individual actions to mitigate the negative effect of domain shift: instead of relying solely on the visual content of a single clip, which may be affected by confusing cues such as unfamiliar backgrounds, occlusions, or user-specific behavior, the model can reason about the relations between the different actions in the sequence to infer the correct action.}
We define an \textit{action sequence} $\mathcal{S}_i$ as the ordered set of $W$ actions centered around action $a_i$, represented by:
\begin{equation}
    \mathcal{S}_i = \{a_{i-W/2}, \dots, a_{i}, \dots, a_{i+W/2}\},
\end{equation}
where each action is a tuple $a_i=(v_i, t_i, y_i, d_i)$ consisting of a \textit{video clip}~$v_i$ of $\mathcal{F}$ frames, a \textit{free-form text narration}~$t_i$ as a textual description of the action, an action label~$y_i$ 
, and $d_i$ identifies one of the $N_s$/$N_t$ source/target domains, respectively. 

We use pre-trained and frozen visual and text features extractors on the video clip $v_i$ and text narration $t_i$ to obtain the visual $\bm{X}^V_i\in\mathbb{R}^{W\times D_V}$ and textual $\bm{X}^T_i\in\mathbb{R}^{W\times D_T}$ features for each action $a_i$ in the sequence.
Here, $W$ represents the length of the sequence, and $D_V$ and $D_T$ correspond to the feature size for the visual and textual input respectively.

To encourage the model to share information between the actions in the sequence, we exploit the self-attention mechanism of transformers~\cite{vaswani2017attention}, which naturally models relations between elements in a sequence.
These are built as a stack of multi-head attention, feed-forward (\texttt{FF}) layers, residual connection, and normalization layers (\texttt{LN}).  

Let $\bm{H}^l$ be the features encoded at the $l$-th layer. Features produced at each layer of the transformer encoder are:
\begin{equation}\label{eq:2}
    \bm{H}^l_{attn}=\texttt{LN}(f_{SA}(\bm{H}^l)+\bm{H}^l),
\end{equation}
\begin{equation}
    \bm{H}^{l+1}=\texttt{LN}(\texttt{FF}(\bm{H}^l_{attn})+\bm{H}^l_{attn}),
\end{equation}
where $f_{SA}(\cdot)$ is the self-attention operator defined as:
\begin{equation}\label{eq:3}
    f_{SA} = \sigma\left(\frac{q(\bm{H}^l) k(\bm{H}^l)}{\sqrt{D}}\right)v(\bm{H}^l),
\end{equation}
where $q$, $k$ and $v$ are learnable projections, 
$D$ is the size of the input features and $\sigma(\cdot)$ is the softmax function.

We feed $\bm{X}^V_i$ to a vanilla transformer encoder~$\texttt{ENC}^\texttt{V}$ 
to aggregate information from all the actions in the sequence. 
As transformers are permutation invariant, we employ positional encoding to preserve information about the order of actions in the sequence. These are learnable vectors $p \in \mathbb{R}^{W \times D_p}$, which are added to the action features to indicate their absolute position inside the sequence. 
Finally, we append a learnable classification token to the end of the sequence, denoted as $\texttt{CLS}$ $\in \mathbb{R}^{D_V}$, as in~\cite{kazakos_temporal2021}. 
The inputs $\bm{X}^V_i$ and $\bm{X}^T_i$ are transformed as follows:

\begin{equation}
\begin{split}
    \bm{Z}_i^V &= \texttt{ENC}^\texttt{V} \left( [\bm{X}^V_i +p_i, \; \texttt{CLS}] \right), \\
    \bm{Z}_i^T &= \mathcal{I} \left( \bm{X}^T_i \right),
\end{split}
\end{equation}
where $\bm{Z}_i^V \in\mathbb{R}^{(W+2)\times D}$ and $\bm{Z}_i^T \in\mathbb{R}^{W\times D}$ are the final output embeddings encoding the relationships between different actions in the sequence and~$\mathcal{I}$ indicates the identity function. This process is illustrated in Figure~\ref{fig:method}.
\subsection{Sequence Reconstruction}
\label{sec:seqrec}
We mask the visual or textual features of the central action in the sequence and learn how to reconstruct it using the 
sorrounding actions. As shown in Section \ref{sec:ablation}, this increases the benefits of using sequences across domains.
Formally, given the visual $\bm{Z}_i^V$ and textual $\bm{Z}_i^T$ features, we zero out the central actions:
\begin{equation}
\begin{split}
    \bm{\bar{Z}}_i^V &= \{z_{i-W/2}^V, \dots, \bar{z}_i^V, \dots, z_{i+W/2}^V\},\\
    \bm{\bar{Z}}_i^T &= \{z_{i-W/2}^T, \dots, \bar{z}_i^T, \dots, z_{i+W/2}^T\},
\end{split}
\end{equation}
where \revised{$\bar{z}_i^V$} and $\bar{z}_i^T$ represent the masked features.
Two separate transformer decoders, $\texttt{DEC}^\texttt{V}$ and $\texttt{DEC}^\texttt{T}$, are trained on $\bm{\bar{Z}}_i^V$ and $\bm{\bar{Z}}_i^T$ to reconstruct the original visual and textual input $\bm{Z}_i^V$ and $\bm{Z}_i^T$ respectively. 
We 
learn to reconstruct the masked visual features with the guidance of the unmasked textual features and vice versa.
To do this, during the decoding process, the masked visual features are combined with the unmasked textual features using a cross-attention mechanism~\cite{vaswani2017attention}. 
The resulting  reconstructed vectors $\bm{\hat{Z}}_i^V$ and $\bm{\hat{Z}}_i^T$ are: 
\begin{equation}\label{eq:4}
    \begin{split}
        \bm{\hat{Z}}_i^V = \texttt{DEC}^\texttt{V}(\bm{\bar{Z}}_i^V, \bm{Z}_i^T) ,\hspace{.5cm} \bm{\hat{Z}}_i^T = \texttt{DEC}^\texttt{T}(\bm{\bar{Z}}_i^T, \bm{Z}_i^V).
    \end{split}
\end{equation}
In the decoding stage, each decoder layer operates similarly to Eq.~\ref{eq:2} and Eq.~\ref{eq:3}, where $f_{SA}$ is replaced by the cross-attention operator $f_{CA}$. At each visual and text decoder layer $\bm{K}^l_V$ and $\bm{K}^l_T$, cross-attention $f_{CA}$ is defined as:
\begin{equation}
    f_{CA}(\bm{K}^l_V, \bm{K}^l_T) = \sigma\left(\frac{q(\bm{K}^l_V) k(\bm{K}^l_T)}{\sqrt{D}}\right)v(\bm{K}^l_V)
\end{equation}
where $q$, $k$ and $v$ are learnable projections of the input features, $\bm{K}^l_V$ and $\bm{K}^l_T$ are features decoded at the $l$-th layer by the visual and textual decoder respectively, $D$ is the size of the input features and $\sigma(\cdot)$ is the softmax function.
The decoders $\texttt{DEC}^\texttt{V}$ and $\texttt{DEC}^\texttt{T}$ are trained by minimizing the L2 distance between the original and the reconstructed visual features ($\mathcal{L}_{rV}$) and the cross-entropy loss between reconstructed textual features and the original text narrations ($\mathcal{L}_{rT}$) respectively. 

\secondrevised{Alternative strategies could be employed to integrate visual and textual modalities in the reconstruction process, for example, by concatenating their embeddings or by using co-attention.
We opt for cross-attention as it can be integrated nicely in \ours architecture by adding cross-attention layers after each decoder layer. This way, each decoder layer implements temporal reasoning over the sequence independently within each modality, while cross-attention allows modalities to peek at each other. This approach effectively integrates cross-modal cues in the reconstruction process, without explicitly mixing the two modalities, which might hurt the unimodal reconstruction objective of \ours.}
\begin{figure}[t]
    \centering
    \includegraphics[width=0.95\columnwidth]{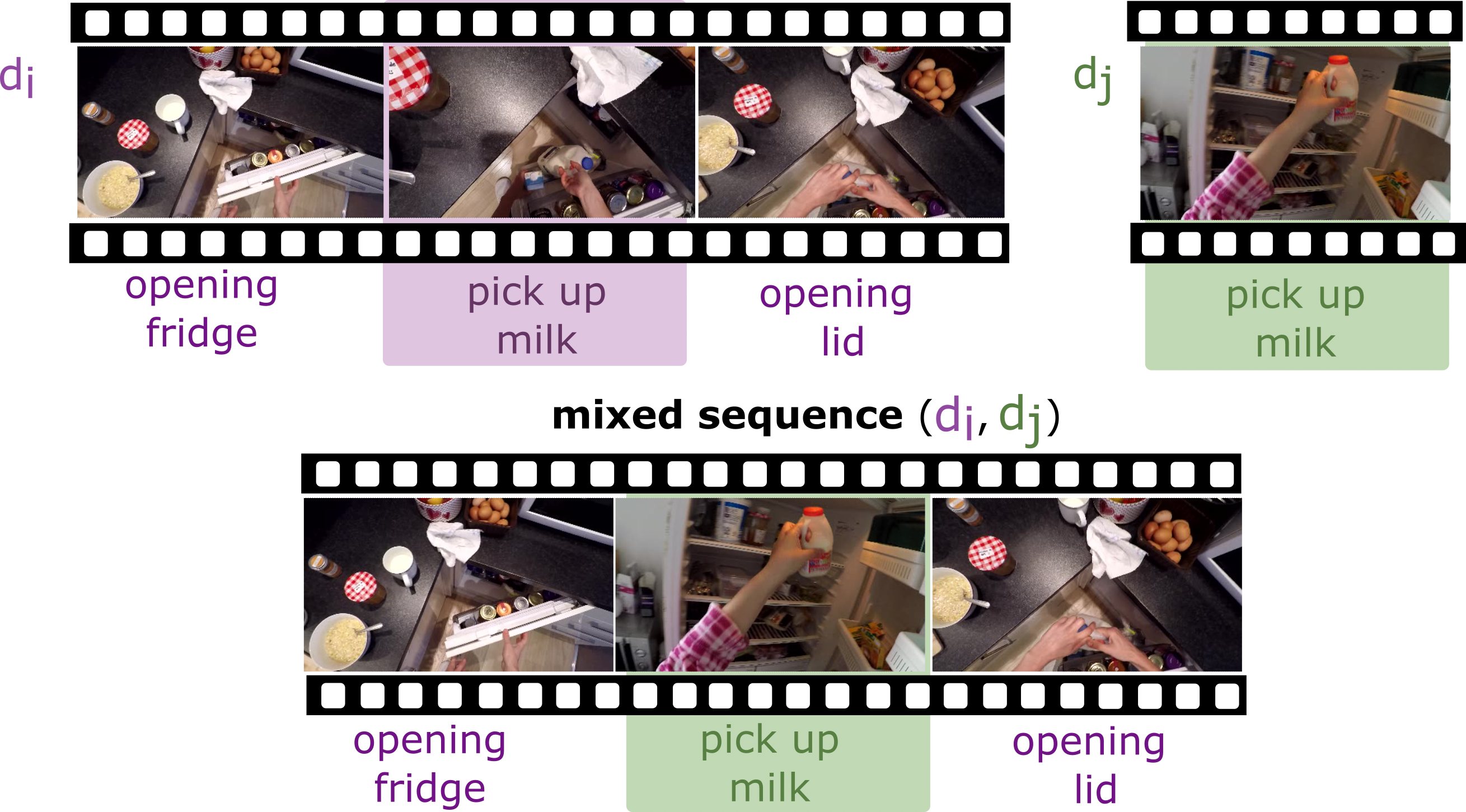}
    \caption{\textbf{SeqMix.} Given a sequence from domain $d_i$, we replace action $a_i$ in the sequence with a different action $a_j$ from another sequence belonging to a different domain $d_j$.}
    \label{fig:seqmix}
\end{figure}
\subsection{SeqMix: Sequence Mix Augmentation}
\label{sec:seqmix}
Sequences $\mathcal{S}_i$ are composed of actions from the same video, thus sharing a similar visual appearance of the environment.
We generate new ‘‘{mixed sequences}" of actions from different source domains to train the network. The rationale behind this approach is to increase the diversity of the training sequences, thereby enhancing the model's ability to generalize to sequences from different domains~\cite{zhang2017mixup}.

During training, given an action sequence $\mathcal{S}_i$, we randomly replace (with probability $p=0.5$) an action $a_i$ in the sequence with a different action $a_j$ from another sequence $\mathcal{S}_j$. The two actions are from different visual domains but share the same action label:
\begin{equation}
 (a_i,a_j): \;\; d_i \neq d_j\;\wedge\; y_i=y_j   
\end{equation}
where $d_i, d_j$ and $y_i,y_j$ are the domain and action labels of the two samples respectively. 
Both $d_i$ and $d_j$ are domains from the source split of the dataset, \ie $d_i,d_j \in \{\mathcal{D}_1^s,\dots,\mathcal{D}_{N_s}^s\}$.
An example of SeqMix is shown in Figure \ref{fig:seqmix}. \\

\subsection{\revised{Training losses}}
For action recognition, we add a classification head, denoted as $h$, which receives the transformed classification token $\bm{Z}^v_{0:1}$ as input, producing the action classification logits $\hat{y}_i$ as output:
\begin{equation}
\hat{y}_i =  h(\bm{Z}^v_{0:1}).
\end{equation}
The classification head is trained using cross-entropy loss on the central action of the sequence ($\mathcal{L}_{C}$) using the corresponding ground truth label $\hat{y}_i$,
\revised{ while the visual and textual reconstruction losses ($\mathcal{L}_{rV}$, $\mathcal{L}_{rT}$) are defined as the mean squared error loss between the reconstructed features $\bm{\hat{Z}}_i^V$ and $\bm{\hat{Z}}_i^T$ and the original unmasked features $\bm{Z}_i^V$ and $\bm{Z}_i^T$, respectively.}
The network is trained jointly using the sum of the classification and reconstruction losses. 
Overall, the training objective of \ours is defined as:
\begin{equation}
\mathcal{L} = \mathcal{L}_{C}+\lambda_{rV}\mathcal{L}_{rV}+\lambda_{rT}\mathcal{L}_{rT}
\end{equation}
where $\lambda_{rV}$ and $\lambda_{rT}$ weight the two reconstruction losses for the visual and textual features.
\subsection{Inference}
The encoder-decoder architecture, trained with text narrations through the reconstruction task, encourages the visual encoder $\texttt{ENC}^V$ to more effectively integrate information from surrounding actions. 
At test time, sequences made of only video clips $v_i$ are processed by the visual encoder $\texttt{ENC}^V$ and classifier $h$. The prediction is performed on the central action and a sliding window mechanism is used for sequential testing. 
Textual annotations are not used at inference time.


\section{Experiments}
\begin{table*}[t]

\centering
\scriptsize
\caption{Experimental results on the \ek~\cite{Damen2018EPICKITCHENS} UDA benchmark, comparing both UDA and DG methods in the \textit{Cross-Domain} setting (target validation set). Models are evaluated in terms of Top-1 and Top-5 Verb, Noun and Action accuracy (\%). For each method, we indicate with \up the improvement over the Source Only. $^\ast$Reproduced results from~\cite{peirone2025egocentric}.}
\renewcommand{\arraystretch}{0.95}
\begin{tabularx}{\textwidth}{Xc|c|ccc|ccc|ccc}

\toprule
\rowcolor{lightgray} & & & \multicolumn{3}{c|}{\textbf{Modalities}} &\multicolumn{3}{c|}{\textbf{Top-1 Accuracy (\%)}} & \multicolumn{3}{c}{\textbf{Top-5 Accuracy (\%)}} \\

\rowcolor{lightgray} \textbf{Method} & \textbf{\revised{Backbone}} & \textbf{Sequence} & \textbf{RGB} & \textbf{Flow} & \textbf{Audio} & \textbf{Verb} & \textbf{Noun} & \textbf{Action} & \textbf{Verb} & \textbf{Noun} & \textbf{Action} \\


\midrule
\rowcolor{lightgray} \multicolumn{12}{c}{\textbf{UDA}}\\
\midrule

Source Only & \revised{TBN-TRN~\cite{kazakos2019epic}} & - & \cmark & \cmark & \cmark &  46.7 & 27.8 & 19.2 & 75.4 & 48.3 & 42.2 \\  
TA3N~\cite{chen2019temporal} & \revised{TBN-TRN~\cite{kazakos2019epic}} & - & \cmark & \cmark & \cmark & 48.5 & 28.9 & 19.6 \scriptsize{(\up+0.4)} & 76.0 & 50.1 & 43.4 \scriptsize{(\up +1.2)}\\ 

\midrule

Source Only & \revised{TBN-TRN~\cite{kazakos2019epic}} & - & \cmark & \cmark & \cmark &  47.1 & 27.4 & 19.0 & 75.3 & 49.4 & 41.8 \\
MM-SADA~\cite{munro2020multi} & \revised{TBN-TRN~\cite{kazakos2019epic}} & - &  \cmark & \cmark & \cmark & 48.4 & 28.3 & 19.3 \scriptsize{(\up +0.3)} & 77.6 & 50.6 & 43.4 \scriptsize{(\up +1.6)} \\  
\midrule

Source Only & \revised{TBN-TRN~\cite{kazakos2019epic}} & - & \cmark & \cmark & \cmark &  47.6 & 28.4 & 19.6 & - & - & -  \\  
CIA~\cite{yang2022interact} & \revised{TBN-TRN~\cite{kazakos2019epic}} & - & \cmark & \cmark & \cmark & 48.3 & 29.5 & 20.3 \scriptsize{(\up +0.7)} & - & - & - \\

\midrule

Source Only & \revised{TBN-TRN~\cite{kazakos2019epic}} & - & \cmark & \cmark & \cmark & 46.7 &  26.7 & 18.2 & 75.3 & 48.4 & 41.3 \\  
RNA~\cite{planamente2024relative} & \revised{TBN-TRN~\cite{kazakos2019epic}} & - & \cmark & \cmark & \cmark & \textbf{50.8} & 29.1 & 20.0 \scriptsize{(\up +1.8)} & 80.8 & 52.1 & 46.0 \scriptsize{(\up +4.7)}\\

\midrule


\rowcolor{lightgray} \multicolumn{12}{c}{\textbf{DG}}\\

\midrule



Source Only & \revised{TBN-TRN~\cite{kazakos2019epic}} & - & \cmark & \cmark & \cmark &  47.2 & 27.4 & 19.0 & 75.3 & 49.4 & 41.8 \\
MM-SADA (SS) \cite{munro2020multi} & \revised{TBN-TRN~\cite{kazakos2019epic}} & - &  \cmark & \cmark & \cmark &  47.8 & 27.9 & 19.2 \scriptsize{(\up +0.2)} & 77.1 & 49.8 & 42.9 \scriptsize{(\up +1.1)}\\

\midrule

Source Only & \revised{TBN-TRN~\cite{kazakos2019epic}} & - & \cmark & \cmark & \cmark & 46.7 &  26.7 & 18.2 & 75.3 & 48.4 & 41.3 \\  
RNA~\cite{planamente2024relative} & \revised{TBN-TRN~\cite{kazakos2019epic}} & - & \cmark & \cmark & \cmark &{50.7}& 27.9 & 19.8 \scriptsize{(\up +1.6)} & 80.6 & 51.3 & 45.3 \scriptsize{(\up +4.0)}\\

\midrule
\revised{
Source Only} & \revised{TBN-TRN~\cite{kazakos2019epic}} & - & \cmark & \cmark & \cmark &  \revised{
49.0 }& \revised{
29.1} & \revised{
19.4} & \revised{
80.7} & \revised{
52.4} & \revised{
45.8} \\
\revised{
CIR~\cite{plizzari2023can}}$^\ast$(No text) & \revised{TBN-TRN~\cite{kazakos2019epic}} & - &  \cmark & \cmark & \cmark & \revised{
 49.4 }& \revised{29.1 }& \revised{
 19.4
\scriptsize{(\up +0.0)}} &\revised{
 80.6} & \revised{
53.4} & \revised{
46.8 \scriptsize{(\up +1.0)}}\\  

\revised{CIR~\cite{plizzari2023can}$^\ast$ }& \revised{TBN-TRN~\cite{kazakos2019epic}} & - &  \cmark & \cmark & \cmark &  \revised{48.8} & \revised{29.0} & \revised{19.4 \scriptsize{(\up +0.0)}} & \revised{81.0} & \revised{53.2 }& \revised{46.9 \scriptsize{(\up +1.1)}}\\ 

\midrule

\revised{
Source Only} & \revised{TBN-TRN~\cite{kazakos2019epic}} & - & \cmark & \cmark & \cmark &  \revised{
49.0} & \revised{
29.1} & \revised{
19.4} & \revised{
80.7} & \revised{
52.4} & \revised{
45.8} \\
\revised{
EgoZAR~\cite{peirone2025egocentric}} & \revised{TBN-TRN~\cite{kazakos2019epic}} & - &  \cmark & \cmark & \cmark & \revised{
 50.0 }& \revised{
29.5 }&\revised{
 20.3
\scriptsize{(\up +0.9)}} &\revised{
 \textbf{81.1}} & \revised{
\textbf{53.6}} & \revised{
\textbf{46.9} \scriptsize{(\up +1.1)}}\\  

\midrule

Source Only & \revised{TBN-TRN~\cite{kazakos2019epic}} & \cmark & \cmark & \xmark  & \cmark & 38.3 & 22.4 & 14.3 & 72.3 & 42.6 & 36.4 \\ 
MTCN \cite{kazakos_temporal2021} & \revised{TBN-TRN~\cite{kazakos2019epic}} & \cmark & \cmark & \xmark  & \cmark & 39.8 & 25.5 & 14.8 \scriptsize{(\up +0.5)} & 75.0 & 46.6 & 40.0 \scriptsize{(\up +3.6)} \\ 

\midrule


Source Only & \revised{TBN-TRN~\cite{kazakos2019epic}} & \cmark & \cmark & \xmark  & \cmark & 38.7 & 23.8 & 14.8 & 75.3 & 45.3 & 38.8\\  
\rowcolor[HTML]{fff2cc} \ours & \revised{TBN-TRN~\cite{kazakos2019epic}} & \cmark & \cmark & \xmark & \cmark  & 41.3 & 26.8 & 16.9 \scriptsize{(\up +2.1)} & 75.8 & 48.7 & 41.5 \scriptsize{(\up +2.7)} \\

\midrule

Source Only & \revised{TBN-TRN~\cite{kazakos2019epic}} & \cmark & \cmark & \cmark & \cmark & 46.4 & 26.6 & 18.2 & 76.8 & 51.9 & 42.2 \\
\ours (No text) & \revised{TBN-TRN~\cite{kazakos2019epic}} & \cmark & \cmark & \cmark & \cmark & 47.5 & 28.9 & 19.5 \scriptsize{(\up +1.3)} & 78.9 & 50.6 & 44.1 \scriptsize{(\up +1.9)} \\
\rowcolor[HTML]{fff2cc} \ours & \revised{TBN-TRN~\cite{kazakos2019epic}} & \cmark & \cmark & \cmark & \cmark & 49.1 & \textbf{29.8} & \textbf{20.6} \scriptsize{(\up +2.4)} & 79.7 & 52.6 & {45.8} \scriptsize{(\up +3.6)} \\

\bottomrule

\end{tabularx}

\label{tab:comparison}
\end{table*}


\subsection{Datasets}
We evaluate \ours on the UDA benchmark of the \ek dataset~\cite{Damen2022RESCALING}, a large collection of human activities recorded in a kitchen environment. \secondrevised{\ek provides over 100 hours of videos recorded in 45 different kitchens.} 
The dataset consists of two splits, \textit{source} and \textit{target}, containing labeled and unlabeled samples respectively, \secondrevised{with videos captured in distinct environments by different participants, thus making the shift between the two splits particularly significant.}
Each visual domain $\mathcal{D}_i$ in the dataset corresponds to a different kitchen in which actions were collected. 
Actions are annotated with \textit{(verb, noun)} pairs from a set of 97 verbs and 300 nouns. Models are evaluated in terms of Top-1 and Top-5 accuracy for Verb, Noun and Action predictions. 
The latter is a combination of the Verb and Noun labels and is used to evaluate the ability of the network to predict both.
%
We also evaluate \ours on the EGTEA~\cite{li2018eye} dataset, which consists of 28 hours of videos with about 10k annotated actions from a set of 106 labels. Videos are captured by 32 subjects over  86 unique sessions, which makes it a valuable testbed to evaluate the better generalization offered by sequence-based reasoning.

\subsection{Setting}
For \ek, we consider two different experimental settings.
In the \textit{Intra-Domain} setting, the train and validation/test samples are from the same visual domains, \ie clips have been recorded in the same kitchens in both splits, whereas, in the \textit{Cross-Domain} setting, they are from different kitchens (\textit{location shift}) or from the same kitchen but recorded after a long temporal interval of several years (\textit{temporal shift}).
In this setting, we further distinguish between UDA methods that use unlabeled target data during training and DG methods that only learn from source data.

\subsection{Implementation details}
\revised{For \ek, we use pre-extracted features using the TBN~\cite{kazakos2019TBN} architecture pre-trained on the source split of the dataset, which are the ones officially provided for that setting~\cite{Damen2018EPICKITCHENS}.
Features are extracted from 25 uniformly sampled clips for each input action and have dimension 1024. 
For EGTEA, we use the SlowFast~\cite{feichtenhofer2019slowfast} pre-trained features released by the authors of MTCN~\cite{kazakos_temporal2021} for a fair comparison with previous methods. 
These features are extracted from 10 clips for each action and have dimension 2304. 
For \ek experiments, we sample 5 clips out of the 25 available, while all 10 clips are used for EGTEA.
A Temporal Relation Network (TRN)~\cite{zhou2018temporal} layer is used to summarize information across different clips for each action. For text, we use the provided narrations from the \ek dataset, which consists of simple verb and noun combinations. }

\revised{We encode the narrations using the BERT pre-trained model~\cite{devlin2018bert}. Textual features have size 768. To align the sizes of the textual and visual features, we project the latter using a fully connected layer with output size 768.
For \ek, we use separate classification tokens for the Verb and Noun labels and apply the cross entropy classification loss to~each.}
\revised{\ours is trained with the SGD optimizer, batch size 32 and learning rate $0.005$. Training lasts 100 and 50 epochs for \ek and EGTEA respectively. Learning rate is decreased by a factor 10 at epochs 50 and 75 for \ek, and at epochs 25 and 38 for EGTEA.}

\subsection{Comparison with state-of-the-art}
In line with previous works~\cite{planamente2024relative,yang2022interact}, we report each method with the corresponding baseline reported in the original paper, indicated as \textit{Source Only}, which allows to compare relative improvements of each method.

\noindent\textbf{Cross-Domain Analysis.}
We compare in Table~\ref{tab:comparison} \ours with previous state-of-the-art methods in UDA 
on the \ek dataset.

For a fair comparison, we include results across all modalities, \ie,  RGB, Optical Flow, and Audio, and results on RGB and Audio only, to fairly compare with MTCN~\cite{kazakos_temporal2021}. 
TA3N~\cite{chen2019temporal}, MM-SADA~\cite{munro2020multi} and CIA~\cite{yang2022interact} use data from the target domain to perform adversarial domain alignment. 
RNA~\cite{planamente2024relative} learns to align features from different modalities and domains. \revised{Additionally, CIR~\cite{plizzari2023can} and EgoZAR~\cite{peirone2025egocentric} can only use source train data, where CIR represents an action as a weighted combination of actions from other scenarios and locations and EgoZAR extracts domain-agnostic representations for activity-centric zones.}
\ours demonstrates better or comparable performance than these methods, in terms of both absolute accuracy and relative improvements compared to the baseline of each method, i.e., the corresponding Source Only. In particular, when considering the combination of verb and noun (Action accuracy), SeqDG achieves 20.6\% accuracy, while the best performing method, 
\revised{EgoZAR}, achieves 20.3\%. 

When looking at relative improvements with respect to the Source Only, SeqDG improves by up to 2.4\%, while the method with the highest improvement in this case, RNA, improves by up to 1.8\%. Overall, we observe that \ours performs best in the Noun and Action metrics, with state-of-the-art Top-1 \revised{accuracy}.
Slightly lower improvements are observed in the Verb metrics, where RNA achieves slightly better results (50.8\% vs 49.1\%). Intuitively, we observe that manipulated objects tend to remain the same between consecutive actions. For example, an apple is first washed, then peeled and finally cut. Differently, verbs may be more diverse and their temporal order may vary, making the reconstruction objective more challenging.
Note that, unlike these methods, \ours does not use the target domain and can also work with only one modality, differently from multi-modal approaches like RNA.

The closest approach to \ours is MTCN~\cite{kazakos_temporal2021}. 
Despite not being designed for Domain Generalization, its sequence-based architecture is similar to ours. 
We compare MTCN and \ours using the same set of modalities, \ie RGB and Audio. The small improvements of MTCN in the \textit{Cross-Domain} setting compared to its baseline (+0.5\%) 
show that naively using information from neighboring actions is not sufficient across different domains.
Instead, \ours is built to explicitly exploit action sequences to improve generalization, which results in consistent improvements over MTCN (14.8\% vs 16.9\%).

Finally, to demonstrate that the improvements of \ours are not merely due to the integration of text in the reconstruction process, we also include a variant of our approach that does not use text entirely (SeqDG (No text)). We observe that even this variant of \ours significantly improves over the baseline (+1.3\%), underlining the importance of exploiting action sequences in a cross-domain scenario.

\begin{table}[t]

\centering
\scriptsize
\caption{Results on the Intra-Domain setting of \ek using RGB and Audio modalities. $^\dagger$Test time predictions filtering using a LM~\cite{kazakos_temporal2021}.}
\renewcommand{\arraystretch}{0.95}
\begin{tabularx}{\columnwidth}{Xc|ccc}
\toprule

\rowcolor{lightgray} &\multicolumn{4}{c}{\textbf{Top-1 Acc. (\%)}}  \\

\rowcolor{lightgray} \textbf{Method} & \revised{Backbone} &  \textbf{Verb} & \textbf{Noun}  & \textbf{Action} \\
\midrule

Source Only & \revised{TBN-TRN~\cite{kazakos2019epic}} &  60.0 & 42.8 & 30.8 \\
MTCN~\cite{kazakos_temporal2021} & \revised{TBN-TRN~\cite{kazakos2019epic}} &   60.6 & 48.2 & 33.0 \\
MTCN$^\dagger$~\cite{kazakos_temporal2021} & \revised{TBN-TRN~\cite{kazakos2019epic}} &   61.5 & 49.3 & 34.3\\

\midrule

Source Only & \revised{TBN-TRN~\cite{kazakos2019epic}} &  60.1 & 42.8 & 31.1 \\ 
\ours & \revised{TBN-TRN~\cite{kazakos2019epic}} &  \textbf{63.9} & 49.1 & 36.1 \\
SeqDG$^\dagger$ & \revised{TBN-TRN~\cite{kazakos2019epic}} &   63.6 & \textbf{49.7} & \textbf{36.3} \\

\bottomrule

\end{tabularx}

\label{tab:comparison_intra}
\end{table}
\begin{table}[t]
    \centering
    \scriptsize
    \caption{Results on the EGTEA dataset in terms of Top-1 accuracy (\%) and Mean Class accuracy (\%).  \revised{$^\dagger$Test time predictions filtering using a LM~\cite{kazakos_temporal2021}.}}
    \renewcommand{\arraystretch}{0.95}
    \begin{tabularx}{\columnwidth}{Xc|cc}
    
        \toprule
        \rowcolor{lightgray} & & \revised{Top-1} & \revised{Mean Class}\\
        \rowcolor{lightgray} \textbf{Method} & \revised{Backbone} & \textbf{Acc. (\%)} & \textbf{Acc. (\%)} \\
        \midrule
        
        Kapidis et al.~\cite{kapidis2019multitask}  & \revised{MFNet~\cite{chen2018multi}} & 68.9 & 60.5 \\ 
        Lu et al.~\cite{lu2019learning}             & \revised{I3D~\cite{carreira2017quo}} & 68.6 & 60.5 \\ 
        SlowFast~\cite{feichtenhofer2019slowfast}   & \revised{SlowFast~\cite{feichtenhofer2019slowfast}} & 70.4 & 61.9 \\ 
        
        Min et al.~\cite{min2021integrating}        & \revised{I3D~\cite{carreira2017quo}} & 68.5 & 62.8 \\ 

        \revised{Shiota et al.~\cite{shiota2024egocentric}} & \revised{SlowFast~\cite{feichtenhofer2019slowfast}} & \revised{69.3} & \revised{ 60.9} \\ 
        
        MTCN~\cite{kazakos_temporal2021}            & \revised{SlowFast~\cite{feichtenhofer2019slowfast}} & 72.5 & 64.8 \\ 
        
        \ours                                       & \revised{SlowFast~\cite{feichtenhofer2019slowfast}} & 74.0 & 66.5 \\ 
        \midrule
    
        MTCN$^\dagger$~\cite{kazakos_temporal2021} & \revised{Slowfast~\cite{feichtenhofer2019slowfast}} & 73.5 & 65.8 \\ 
        SeqDG$^\dagger$ & \revised{Slowfast~\cite{feichtenhofer2019slowfast}} & \textbf{74.1} & \textbf{66.9} \\ 
    
    \bottomrule
    \end{tabularx}

    \label{tab:egtea}
\end{table}
\begin{table}[t] 
    \centering
    \scriptsize
    
    \caption{Ablation study on the different components of \ours on \ek both on RGB-only and all modalities (RGB, Flow, Audio).}
    
    \setlength{\tabcolsep}{4pt}
    \renewcommand{\arraystretch}{0.95}
    \begin{tabular}{c|cccc|ccc}
    \toprule
        \rowcolor{lightgray} &&&&&\multicolumn{3}{c}{\textbf{Top-1 Acc. (\%)}}  \\
        \rowcolor{lightgray} \textbf{\revised{Setting}} & \textbf{Seq.} & \textbf{SeqMix} & \textbf{$\mathcal{L}_{rV}$} & \textbf{$\mathcal{L}_{rT}$} & \textbf{Verb}  & \textbf{Noun}  & \textbf{Action} \\
        \midrule
        
        \revised{RGB-only} & - & - & - & -  & 33.5 & 21.6 & 11.6 \\  
        \revised{RGB, Flow, Audio} & - & - & - & -  & \revised{46.4} & \revised{26.6} & \revised{18.2} \\  
        \midrule
        \revised{RGB-only}& \cmark & - & - & -  & 32.9 & 22.7 & 12.0 \\
        \revised{RGB, Flow, Audio}& \cmark & - & - & -  & \revised{46.6} & \revised{28.1} & \revised{19.3} \\
        \midrule
        \revised{RGB-only}& \cmark & \cmark & - & -  & 33.9 & 23.2 & 12.1 \\
        \revised{RGB, Flow, Audio}& \cmark & \cmark & - & -  & \revised{47.5} & \revised{29.1} & \revised{19.7} \\
        \midrule
        \revised{RGB-only}& \cmark & \cmark & \cmark & -  & 33.5 & 23.7 & 12.4 \\ 
        \revised{RGB, Flow, Audio}& \cmark & \cmark & \cmark & -  & \revised{47.9} & \revised{28.6} & \revised{19.8} \\ 
        \midrule
        \revised{RGB-only}& \cmark & \cmark & - & \cmark  & 34.2 & 23.6 & 12.2 \\ 
        \revised{RGB, Flow, Audio}& \cmark & \cmark & - & \cmark  & \revised{47.7} & \revised{29.2} & \revised{19.8} \\ 
        \midrule
        \revised{RGB-only}& \cmark & \cmark & \cmark & \cmark  & 34.3 & 24.2 & 12.8 \\ 
        \revised{RGB, Flow, Audio}& \cmark & \cmark & \cmark & \cmark  & \textbf{\revised{49.1}} & \textbf{\revised{29.8}} & \textbf{\revised{20.6}} \\ 
        
        \bottomrule
    \end{tabular} 
    \label{tab:ablation}
\end{table}
\begin{table}[t]
    \centering
    \scriptsize
    \caption{Multi-modal performance of \ours on \ek.}
    \begin{tabularx}{\columnwidth}{X|ccc|ccc}
    \toprule
        \rowcolor{lightgray} & \multicolumn{3}{c|}{\textbf{Modalities}} & \multicolumn{3}{c}{\textbf{Top-1 Acc. (\%)}}  \\
        \rowcolor{lightgray} \textbf{Setting} & \textbf{RGB} & \textbf{Flow} & \textbf{Audio} & \textbf{Verb} & \textbf{Noun} & \textbf{Action} \\
        \midrule
        Source Only & \cmark & - & - & 33.5 & 21.6 & 11.6 \\
        \ours & \cmark & - & - & 34.3 & 24.2 & 12.8 \\
        \midrule
        Source Only & \cmark & \cmark & - &  42.1 & 24.3 & 15.3 \\
        \ours & \cmark & \cmark & -  &          45.0 &         28.4 &          18.0 \\
        \midrule

        Source Only & \cmark & - & \cmark & 38.7 & 23.8 & 14.8 \\
        \ours & \cmark & - & \cmark  &          41.3 &         26.8 &          16.9 \\  
        \midrule

        Source Only & \cmark & \cmark & \cmark & 46.4 & 26.6 & 18.2 \\
        \ours & \cmark & \cmark & \cmark  &          \textbf{49.1} &         \textbf{29.8} &          \textbf{20.6} \\ 
        \bottomrule
    \end{tabularx}

    \label{tab:modalities}
    
\end{table}

\begin{table}[!ht]
    \centering
    \scriptsize
    \caption{Comparison of different LMs for \ours \revised{(RGB-only)}.}
    \begin{tabularx}{\columnwidth}{X|ccc}
    \toprule
    \rowcolor{lightgray} &\multicolumn{3}{c}{\textbf{Top-1 Acc. (\%)}}  \\
    \rowcolor{lightgray} \textbf{Language Model} & \textbf{Verb}  & \textbf{Noun}  & \textbf{Action} \\
    \midrule
    CLIP~\cite{radford2021learning} & 34.4 & 23.2 & 12.3 \\ 
    MiniLM~\cite{wang2020minilm}    & \textbf{34.8} & 23.6 & 12.4 \\ 
    BERT~\cite{devlin2018bert}      & 34.3 & \textbf{24.2} & \textbf{12.8} \\ 
    \bottomrule
    \end{tabularx}

    \label{tab:lm_comparison}
\end{table}
\begin{table}[!ht]
\centering
\scriptsize
\caption{\secondrevised{Comparison of different visual backbones for \ours on \ek~\cite{Damen2018EPICKITCHENS} using RGB and Flow modalities.}}
\begin{tabularx}{\columnwidth}{X|c|ccc}

\toprule

\rowcolor{lightgray}  & & \multicolumn{3}{c}{\secondrevised{\textbf{Top-1 Accuracy (\%)}}} \\

\rowcolor{lightgray} \secondrevised{\textbf{Method}} & \secondrevised{\textbf{Backbone}} & \secondrevised{\textbf{Verb}} & \secondrevised{\textbf{Noun}} & \secondrevised{\textbf{Action}} \\
\midrule
\secondrevised{Source Only }& \secondrevised{TSN-TRN~\cite{wang2016temporal} }&  \secondrevised{44.0 }& \secondrevised{25.7 }& \secondrevised{15.8 }\\
\secondrevised{\ours} & \secondrevised{TSN-TRN~\cite{wang2016temporal}} &  \secondrevised{46.0} & \secondrevised{27.4 }& \secondrevised{17.6 \scriptsize{(\up +1.8)} }\\
\midrule
\secondrevised{Source Only} & \secondrevised{I3D-TRN~\cite{carreira2017quo}} &  \secondrevised{46.9} & \secondrevised{25.8 }& \secondrevised{16.6 }\\ 
\secondrevised{\ours} & \secondrevised{I3D-TRN~\cite{carreira2017quo}} &  \secondrevised{\textbf{48.7}} & \secondrevised{\textbf{28.7}} & \secondrevised{19.3 \scriptsize{(\up +2.7)}} \\ 
\midrule
\secondrevised{Source Only }& \secondrevised{TSM-TRN~\cite{lin2019tsm} }&  \secondrevised{45.7 }&  \secondrevised{26.5 }& \secondrevised{17.0 }\\  
\secondrevised{\ours} & \secondrevised{TSM-TRN~\cite{lin2019tsm}} &  \secondrevised{48.5 }&  \secondrevised{28.3 }& \secondrevised{\textbf{19.4}  \scriptsize{(\up +2.4)}}\\  
\midrule
\secondrevised{Source Only} & \secondrevised{TBN-TRN~\cite{kazakos2019epic}} & \secondrevised{42.1 }& \secondrevised{24.3} & \secondrevised{15.3} \\
\secondrevised{\ours} &\secondrevised{ TBN-TRN~\cite{kazakos2019epic}} & \secondrevised{45.0} & \secondrevised{28.4} & \secondrevised{18.0  \scriptsize{(\up +2.7)}}\\

\bottomrule
\end{tabularx}
\label{tab:backbone}
\end{table}

\noindent\textbf{Intra-Domain Analysis.}
In Table~\ref{tab:comparison_intra} we also evaluate the performance of \ours in the intra-domain setting of \ek, \ie the visual domains are shared between the source and target domains.
This setting better highlights \ours's improvements in exploiting action sequences for action recognition, even when domain shift is not present.
\revised{Due to the similarity that exists between MTCN and \ours in using sequence,} we 
\revised{compare \ours} with MTCN using the same modalities, \ie RGB and Audio. 
Additionally, MTCN uses \revised{
Language Model (LM)} at test time to filter predictions by looking for the most likely sequence using the top-k predictions for each action. 
To ensure a fair comparison, we adopt the same approach at test-time for both MTCN and \ours, which we refer to as MTCN$^\dagger$ and SeqDG$^\dagger$. 
Overall, \ours outperforms MTCN by 3\revised{.1}\% when not using text information at test-time, and by \revised{
2\%} when using it.
We observe that the effect of filtering predictions at test time using text is more limited with SeqDG$^\dagger$ compared to MTCN$^\dagger$, indicating that the network has already learned to effectively use text in the reconstruction process. 
%


We report in Table~\ref{tab:egtea} results of \ours 
on the EGTEA dataset. 
\ours outperforms all existing methods, achieving SOTA results on both Top-1 accuracy (\%) and Mean Class accuracy (\%).
Also in this setting, we observe limited benefits by filtering predictions at test time using text, which further confirms \ours learns to effectively exploit text during its training process. Most notably, \ours without filtering outperforms MTCN with predictions filtering.
\subsection{Ablations}
\label{sec:ablation}
\noindent\textbf{\ours components.}
In Table~\ref{tab:ablation}, we illustrate the impact of the different components of \ours. 
Specifically, we compare the use of sequences and individual actions for action recognition, the effect of \oursmix, and the two cross-modal reconstruction losses, $\mathcal{L}_{rV}$ and $\mathcal{L}_{rT}$. 
We find that using sequences 
improves results by \revised{1.1\% using RGB-only, and 1.5\% with all modalities (RGB, Flow, and Audio)} on Nouns and \revised{0.4\% using RGB-only, and 1.1\% with all modalities (RGB, Flow, and Audio)} on Action. 
\oursmix further improves results for both Verbs and Nouns by up to 1\%. 
Mixing actions from different sequences is a key component of \ours to improve generalization, as it encourages the encoder to focus more on the semantics of the actions rather than their visual appearance, which is more dependent on the domain.
Using the reconstruction losses on textual features ($\mathcal{L}_{rT}$) and visual features ($\mathcal{L}_{rV}$) further improves performance. 

\smallskip
\noindent \textbf{Ablation on different modalities.}
Other modalities, such as Audio and Optical flow, may also experience domain shift in forms other than visual appearance. For example, objects' shape and material may affect how interactions are captured by optical flow and audio, respectively.
We show the multi-modal performance of \ours in Table~\ref{tab:modalities}, observing even larger performance improvements compared to the single modality setting. This not only indicates that our approach can be easily extended to other modalities, but also shows that the complementary information brought by them helps in the reconstruction process.

\smallskip
\noindent\textbf{Language Model.}
We evaluate different Language Models for \ours \revised{(RGB only)} to extract the textual features used in the reconstruction process in Table~\ref{tab:lm_comparison}. 
Despite the LMs having very different model sizes, we observe comparable results for Verb and Noun accuracy. We remark that text is only used to guide the reconstruction, 
but it is not part of the inference process.

\smallskip
\noindent\secondrevised{\textbf{Backbone.} In Table~\ref{tab:backbone}, we present the results of \ours on TSN~\cite{wang2016temporal}, I3D~\cite{carreira2017quo}, TSM~\cite{lin2019tsm}, and TBN~\cite{kazakos2019epic},
using the same subset of modalities (RGB, Flow) available for all models to ensure a fair comparison. For each backbone, we compare the performance of the Source Only baseline with \ours. Critically, across \textit{all} backbone architectures, \ours consistently outperforms the Source Only baseline, highlighting that the method is effective and agnostic to the specific visual encoder used.} 

\smallskip
\noindent\textbf{Sequence length.} 
\begin{table}[t]
\centering
\scriptsize
\caption{Ablation study of different sequence lengths on Top-1 accuracy (\%) for action classification of \ek (RGB-only) using \ours.}
\begin{tabularx}{\columnwidth}{X|c|ccc}
\toprule
\rowcolor{lightgray}  & & \multicolumn{3}{c}{\textbf{Top-1 Accuracy (\%)}} \\
\rowcolor{lightgray} \textbf{Method} & \textbf{Seq. Length} & \textbf{Verb} & \textbf{Noun} & \textbf{Action} \\
\midrule
Source Only & 1 & 33.5 & 21.6 & 11.6 \\
\ours & 3 & 34.0 \scriptsize{(\up +0.5)} & 23.2 \scriptsize{(\up +1.6)} & 12.1 \scriptsize{(\up +0.5)}\\
\ours & 5 & 34.3 \scriptsize{(\up +0.8)} & \textbf{24.2} \scriptsize{(\up \textbf{+2.6})} & \textbf{12.8} \scriptsize{(\up \textbf{+1.2})}\\
\ours & 7 & \textbf{34.8} \scriptsize{(\up \textbf{+1.3})} & 23.3 \scriptsize{(\up +1.7)} & 12.6 \scriptsize{(\up +1.0)}\\
\bottomrule
\end{tabularx}
\label{tab:seq_len}
\end{table}
 \begin{figure}[!t]
     \centering
    \includegraphics[width=0.9\columnwidth]{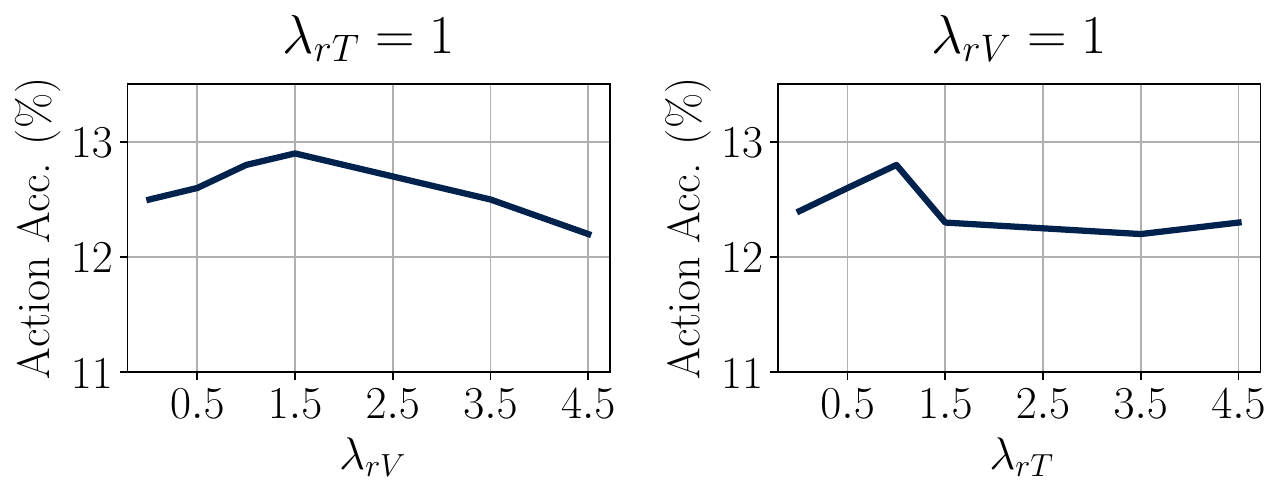}

     \caption{Parameter analysis of the weights associated with the visual and textual reconstruction losses of \ours on \ek (RGB-only).}
     \label{fig:hyperp}
 \end{figure}

In Table~\ref{tab:seq_len}, we analyze the performance of \ours with different sequence lengths ($W)$. We observe the best performance with $W=5$. With shorter sequences, the model receives limited temporal information from both past and future actions. Conversely, increasing the sequence length beyond a certain point may not consistently improve performance, as it can introduce misleading actions that are weakly related to the target action and the overall sequence context, making them less useful for sequence reconstruction (SeqRec).

\smallskip
\noindent\textbf{\revised{Variation on hyper-parameters}} 
Figure~\ref{fig:hyperp} shows the effect of different loss weights for the visual and textual reconstruction losses, when one of two is fixed and the other is varying. Overall, we observe the best performance when the losses weight is equal to one ($\lambda_{rV}$=$\lambda_{rT}$=1.0). These results can be interpreted as higher weights for the SeqRec components, leading to less focus on classification, while the lower values for $\lambda_{rV}$ and $\lambda_{rT}$ reduce the knowledge distillation capability of \ours.

\section{Conclusion and Future Works}
In this work, we propose a method that leverages sequence consistency across diverse environments to improve generalization in egocentric action recognition tasks. We validate the effectiveness of our approach through extensive experiments on widely-used egocentric action recognition datasets, achieving state-of-the-art performance.
Future work could focus on identifying action patterns of varying lengths that generalize well across different scenarios while distinguishing user-specific behaviors. This would enhance our model's adaptability to more dynamic environments.
\bibliographystyle{IEEEtran}
\bibliography{bib_file_new}

\end{document}